\documentclass{article}
\usepackage{spconf,amsmath,graphicx}

\usepackage{times}
\usepackage{latexsym}
\usepackage[T1]{fontenc}
\usepackage[utf8]{inputenc}
\usepackage{microtype}
\usepackage{inconsolata}
\usepackage{multirow}
\usepackage{graphicx}
\usepackage{booktabs}
\usepackage{amsmath}
\usepackage{CJKutf8}
\usepackage{array}
\usepackage{float}
\usepackage{url}
\usepackage{xcolor}
\usepackage{hyperref}

\usepackage{xpinyin}
\usepackage{enumerate}
\usepackage{subcaption}
\usepackage{amsmath,amssymb,amsthm, amsfonts,color}
\title{Rethinking the Reasonability of the Test Set \\ for Simultaneous Machine Translation}
\def\star{$^*$}

\name{Author(s) Name(s)}
\address{Author Affiliation(s)}
\name{Mengge Liu~$^{1}$$^{\dagger}$\thanks{$^{\dagger}$Equal Contribution.}$^{\ddagger}$\thanks{$^{\ddagger}$The work was done during the author’s internship at Xiaomi.}\ \  Wen Zhang~$^{2}$$^{\dagger}$\ \  Xiang Li~$^{2}$\ \  Jian Luan~$^{2}$\ \  Bin Wang~$^{2}$\ \  Yuhang Guo~$^{1}$\star\thanks{$^*$Corresponding Author.}\ \  Shuoying Chen~$^{1}$}
\address{$^{1}$Beijing Institute of Technology, Beijing, China \qquad $^{2}$Xiaomi AI Lab, Beijing, China}
\begin{document}
%\ninept
%
\maketitle
\begin{abstract}
Simultaneous machine translation~(SimulMT) models start translation before the end of the source sentence, making the translation monotonically aligned with the source sentence. However, the general full-sentence translation test set is acquired by offline translation of the entire source sentence, which is not designed for SimulMT evaluation, making us rethink whether this will underestimate the performance of SimulMT models. In this paper, we manually annotate a monotonic test set based on the MuST-C English-Chinese test set, denoted as SiMuST-C. Our human evaluation confirms the acceptability of our annotated test set.
Evaluations on three different SimulMT models verify that the underestimation problem can be alleviated on our test set.
Further experiments show that finetuning on an automatically extracted monotonic training set improves SimulMT models by up to $3$ BLEU points.
% 
% But when evaluated by general offline test set
% 脚注放到正文
% Simultaneous Machine Translation (SimulMT) models show dramatical performance drop with the decrease of latency or flicker, when evaluated on a general full sentence translation test set. This naturally leads us to a doubt: whether SimulMT models really perform poorly, or the evaluation result on this test set cannot truly reflect the preformance of SimulMT models.
% In this paper, we observe that the general full sentence translation
% test set actually underestimates the ability of the SimulMT model, especially with low latency or few flicker requirements.
% We manually annotate a monotonic test set~\footnote{Full test set will be released if the paper is accepted.} based on the MuST-C English-Chinese test set.
% The human evaluation confirms that the acceptability of our annotated test set is comparable to the original test set.
% Experiments show that finetuning on a monotonic dataset automatically extracted from the training set improves SimulMT models by up to $3$ BLEU points on our annotated test set.
\end{abstract}
\begin{keywords}
Machine Translation, Simultaneous Machine Translation Evaluation
\end{keywords}\vspace{-0.2cm}
\section{Introduction}
\label{sec:intro}
\vspace{-0.2cm}
% 介绍同传的相关工作，引入问题，整句测试集评估同传模型是否公平
Recently, remarkable progress has been made by simultaneous machine translation~(SimulMT) models~\cite{cho2016can,gu2017learning}, consisting of streaming translation models~\cite{ma2018stacl,ma2020monotonic} that do not revise translations and re-translation models~\cite{niehues2016dynamic, arivazhagan2020re_icassp} with revision. 
Streaming translation models either adopt fixed policies~\cite{ma2018stacl, elbayad2020efficient, yarmohammadi2013incremental} or adaptive policies~\cite{cho2016can, gu2017learning, ma2020monotonic, zhang2022gaussian, miao-etal-2021-generative} to find the READ-WRITE paths and need to balance translation quality and latency. 
Re-translation models re-translate each successive source prefix to revise previous partial translations, requiring careful control of the flicker in the translation~\cite{niehues18_interspeech, arivazhagan2020re_iwslt}. 
However, there is a thought-provoking phenomenon. Most SimulMT models are evaluated on the general full-sentence translation test set, which is acquired by translating the full source sentence offline. Yet the SimulMT models must generate translations without reading the full source sentences. This makes us wonder: is it reasonable to evaluate the performance of SimulMT models with the general full-sentence translation test set? 
%is the SimulMT model really bad? Or is the test set insufficient to reflect the performance of the model?
% 保留第一个现象，bleu下降剧烈不作为动机，动机是原始测试集没有未同传设计，存在不合适；
%This leads us to become suspicious of the reason, is our SimulMT models not good enough, or our evaluation is not fair enough. 

% 分析同传模型评估存在的问题，对比人工和自动质量评估，说明整句翻译的参考译文低估了同传译文质量
To explore this question, we compare the automatic and human evaluation results of the Wait-$k$ and Re-trans models on the MuST-C test set. 
In Table~\ref{table_bleu_drop}, the translation quality of the Wait-$k$ and Re-trans models degrades rapidly as latency and flicker decrease, and the BLEU scores of both models with low latency and few flickers are $24.2$\% and $23.4$\% lower than those with high latency and many flickers, respectively.
Note that $\Delta$ stands for the quality drop rate, lower AL~\cite{ma2018stacl} value means lower latency, lower NE value~\cite{arivazhagan2020re_icassp} means fewer flickers, and AP is human acceptability~\cite{national1966language,castilho2018approaches}.
A total of $200$ sentences are randomly sampled from the test set for human evaluation\footnote{The evaluator has extensive experience and qualification with TEM-8~(Test for English Majors-Band 8).}.
Surprisingly, we find that both for the Wait-$k$ and Re-trans models, the quality drop rates in human scoring is only $13$\textasciitilde$14$\%, which is much lower than those of BLEU scores.
Therefore, the general full-sentence translation test set indeed underestimates the ability of the SimulMT model.
%Besides,  Online translation always need low-latency or few-flicker, then about $23\%$ descent is observed in Table~\ref{table_bleu_drop}, at low-latency for Wait-k model or few-flicker for Re-translation model. Intuitively, online translation relay on limited the source imformation when transalting each source prefix, so the quality decrease is easy to understand. But the dismatch between the online hypotheses and offline reference can not be ignored, some acceptable hypotheses may have vary low score because the word order is different from the offline reference. As the human evaluation results showed in Table~\ref{table_bleu_drop}, for each inference setting we select the same 200 samples for human evaluation, and the quality decrease is not as big as BLEU score. 

\begin{table}
\centering
\renewcommand\arraystretch{1.1}
{\small
\begin{tabular}{c|c|c|c|c}
\toprule
\multirow{2}*{\bf Model} & \multicolumn{2}{c|}{\bf BLEU}    &   \multicolumn{2}{c}{\bf AP} \\
\cline{2-5}
 & \textbf{Score} & $\Delta$(\%) & \textbf{Score} & $\Delta$(\%)  \\
\hline
Wait-$k$~(AL=$6.08$) & $24.1$ & \multirow{2}*{-$24.2$\%} & $78$\% & \multirow{2}*{-$14.1$\%} \\
Wait-$k$~(AL=$1.09$) & $18.2$ & ~ & $67$\% & ~ \\
% \hline
% \textbf{Model} & \textbf{NE$\downarrow$} & \textbf{BLEU$\uparrow$} & \textbf{QD} & \textbf{AP$\uparrow$} & \textbf{QD}  \\
\hline
Re-trans~(NE=$1.00$) & $25.2$ & \multirow{2}*{-$23.4$\%} & $75$\% & \multirow{2}*{-$13.3$\%} \\
Re-trans~(NE=$0.09$) & $19.3$ & ~ & $65$\% & ~ \\
% Retrans & \tabincell{l}{1\\0.09} & \tabincell{l}{25.2\\19.3} & 23.4\% & \tabincell{l}{xx\\xx} & yy \\
\bottomrule
\end{tabular}
}
\vspace{-0.3cm}
\caption{\label{table_bleu_drop}
\small{Comparison between automatic and human evaluation.}
%  AA is metric about monotonicity, lower means more monotonic. AL is a metric about latency, smaller means lower latency.
}\vspace{-0.7cm}
\end{table}

% 分析低估问题原因，测试集单调性不足，引出标注新测试集的需求
% real-world测试集存在漏译，引出我们新的标注方案，以及微调方案
Intuitively, SimulMT models usually generate monotonic translations due to limited source information. However, the long-distance reordering in the general full-sentence translation test set leads to the problem that the test set underestimates the SimulMT model.
~\cite{chen2020improving,chang2022anticipation} show that the monotonic data and the monotonic training method could improve translation quality at low latency.
~\cite{zhang2021bstc,Machcek2021LostII,zhao2021not} collect real-world interpretation data, which have serious omission because the interpretation task is extremely challenging and exhausting for human.
% ~\cite{zhang2021bstc} create a Chinese-English interpretation test set by annotating Chinese speech with multi-interpreters, but the interpretation has low acceptability because of serious omission. 
% ~\cite{machavcek2021lost} release an English-Czech interpretation corpus and find serious information loss of human interpretation.
% ~\cite{zhao2021not} collect 323-hour German-English interpretation videos from European Parliament Plenary and finally get $1090$ high-quality samples for the test set, showing the difficulty of interpretation data collection.
% the interpretation task is extremely challenging and exhausting for human
% 我们提出新的想法，通过流式文本模拟同传场景，实现高质量的同传译文标注
To this end, we devise a new annotation method performed on text streams, which has no limitation in time or memory for annotators\footnote{The annotators have extensive experience and qualification with TEM-8.}. Our annotation method is applied to the MuST-C~\cite{di2019must,cattoni2021must} English-Chinese test set\footnote{SiMuST-C is available at \url{https://github.com/XiaoMi/SiMuST-C}.}. 
Comparative experiments on three different SimulMT models show that the underestimation problem can be alleviated on our annotated test set.
Moreover, finetuning on a monotonic dataset automatically extracted from the training set improves SimulMT models by up to $3$ BLEU points on our annotated test set.
\section{Method}
\vspace{-0.3cm}
%\item {\bf Monotonic}: Sentence pairs with fewer reordering phenomenon are more~\emph{monotonic}. The more monotonic a sentence pair is, the more it conforms to the style of simultaneous interpretation;
\subsection{Human Annotation}
\vspace{-0.2cm}
% 这里详细介绍我们设计的标注工具
%In this section, we elaborate on how to annotate monotonic translations.
Our annotation is performed on text streams. Initially, no words in the source sentence are exposed to the annotator. The annotater starts with reading the first source word, then he/she chooses either the READ or WRITE action per step. READ means the annotator reads the next source word, and WRITE means the annotator translates and outputs a target word. Once the full source sentence has been read, the annotator finishes the current sentence. An example is given in Table~\ref{table-annoation-example}, the source and target streams are recorded during annotation. We also make an agreement with annotators that the target words that have already been written cannot be revised.\vspace{-0.2cm}
\begin{CJK*}{UTF8}{gbsn}
\begin{table}[!h]
\centering
\renewcommand\arraystretch{1.2}
{\footnotesize
\begin{tabular}{l|l|l}
\toprule
\textbf{Source Streams} & \textbf{Target Streams} & \textbf{Actions}  \\
\hline
And &  & R\big(And\big)  \\
And this & $\overset{\mbox{this}}{\mbox{这}}$ & R\big(this\big) W\big($\overset{\mbox{this}}{\mbox{这}}$\big)  \\
And this made & $\overset{\mbox{this}}{\mbox{这}}$\ \  $\overset{\mbox{make}}{\mbox{使}}$ & R\big(made\big) W\big($\overset{\mbox{make}}{\mbox{使}}$\big)  \\
And this made me & $\overset{\mbox{this}}{\mbox{这}}$\ \  $\overset{\mbox{make}}{\mbox{使}}$ \ \ $\overset{\mbox{me}}{\mbox{我}}$  & R\big(me\big) W\big($\overset{\mbox{me}}{\mbox{我}}$\big)  \\
And this made me sad & $\overset{\mbox{this}}{\mbox{这}}$\ \  $\overset{\mbox{make}}{\mbox{使}}$ \ \ $\overset{\mbox{me}}{\mbox{我}}$ $\overset{\mbox{sad}}{\mbox{难过}}$  & R\big(sad\big) W\big($\overset{\mbox{sad}}{\mbox{难过}}$\big)  \\
\bottomrule
\end{tabular}
}
\vspace{-0.2cm}
\caption{\label{table-annoation-example}
\small{An example of streaming annotation. R and W represent the READ and WRITE actions performed by annotators, respectively. The contents in parentheses indicate what annotators read and write.}\vspace{-0.5cm}
%  AA is metric about monotonicity, lower means more monotonic. AL is a metric about latency, smaller means lower latency.
}
\vspace{-0.2cm}
\end{table}
\end{CJK*}
% 拼音改直译

% Based on the English sentences in the MuST-C English-Chinese test set, annotators output monotonic Chinese translation according to our annotation method. % As shown in Table~\ref{table-annoation-example}, the intermediate target streams are all recorded, so the intermediate output of SimulMT models can be evaluated by calculating the BLEU score based on the target streams.

\vspace{-0.2cm}
\subsection{Automatic Extraction}
% 这里详细介绍从训练集中自动抽取finetune集的过程，以及如何finetune同传模型的
Drawing on $AR_k$ defined by~\cite{chen2020improving}, we design a metric to measure the monotonicity of parallel sentence pairs.
Given a sentence pair $X=[x_1, x_2, ..., x_{i},...]$ and $Y=[y_1, y_2, ..., y_{j},...]$, we use the tool~\texttt{SimAlign}~\cite{jalili-sabet-etal-2020-simalign} to calculate the word alignment $As$.
% ~\footnote{A high-quality word alignment tool based on pretrained models:~\url{https://github.com/cisnlp/simalign}}
The presence of $(x_i, y_j)$ in $As$ means that the $i^{th}$ word in the source sentence is aligned with the $j^{th}$ word in the target sentence.
$i-j>0$ indicates that anticipation~\cite{ma2018stacl} occurs, which means that the $j^{th}$ target word is aligned with the source word that has not yet been seen. Assume $n=|As|$, Average Anticipation~(AA) is computed:
\vspace{-0.2cm}
\begin{small}
\begin{equation} \label{AA-AR}
    \begin{split}
        AA& = \frac{1}{n} \sum_{\tiny{(x_i,y_j)} \in \tiny{As}} \max\left(i-j, 0\right) \\
    \end{split}
    \vspace{-0.15cm}
\end{equation}
\end{small}

We first calculate the AA score of each sentence pair in the training set and then select sentence pairs with AA scores of $0$. We believe that these sentence pairs are relatively monotonic and do not contain long-distance reordering, which is used to finetune the SimulMT models.
\vspace{-0.3cm}
\section{Experiments}
\vspace{-0.3cm}
\label{sec:experiments}
% We first introduce datasets and mainstream SimulMT models, then describe several metrics for evaluating the quality of models and datasets, and finally explore the different performances of models on original and annotated test sets.
\subsection{Datasets}
\label{subsec:datasets}
\vspace{-0.2cm}
% 详细介绍我们使用的训练集和测试集（包括非同传风格的测试集和同传风格的测试集）
% Must-c enzh v2 数据集，介绍下数据集的来源和数量，数据预处理方案（简要）
% 标注的同传风格测试集
% 抽取的同传风格训练集
We use the English-Chinese dataset from MuST-C release v2.0\footnote{\url{https://ict.fbk.eu/must-c-release-v2-0/}}, where the training and development sets consist of $358,853$ and $1,349$ sentence pairs, respectively. The original test set tst-COMMON contains $2,841$ sentence pairs, denoted as~\texttt{test-orig}, and the reference is called OrigRef. Human annotation is performed on the source of~\texttt{test-orig} to build a monotonic test set called~\texttt{test-mono}, and the reference is marked as MonoRef.
Examples in Table~\ref{table:annotation_example} show the difference between OrigRef and MonoRef.
~\texttt{Sacremoses}\footnote{\url{https://github.com/alvations/sacremoses}} and~\texttt{Jieba}\footnote{\url{https://github.com/fxsjy/jieba}} are employed for English tokenization and Chinese word segmentation.
Byte pair encoding~\cite{sennrich-etal-2016-neural} is applied with $32$k operations. 
For the first example, the word ``worldwide'' is translated at the end in the MonoRef, which is more consistent with the word order in the source sentence compared to the OrigRef, so the MonoRef has better monotonicity. 
In the second example, the OrigRef is actually obtained by segmenting the document-level translation into sentences, so translations may depend on the context, such as ``animal,'' whereas the MonoRef is a sentence-level translation, which can only translate information in the source sentence and is more in line with the prediction of machine translation models.
%In Example.2, as the OrigRef is first got by plain text and then aligned into sentences, it may rely on more context information, but our MonoRef is annotated by sentence and more loyal to the source sentence.
% 结合case，说明单句翻译，和机器翻译的输入输出模式更加接近
% The test sets with original reference~(OrigRef) and annotated monotonic reference(MonoRef) are labeled as and~\texttt{test-mono}.
% After filtering, $340$k sentence pairs are reserved for training.
%For the training set, we remove duplicate sentence pairs and discard sentence pairs whose source sentence or target sentence has more than $200$ tokens.
%Sentence pairs with low alignment scores\footnote{The alignment score is calculated by~\texttt{fast\_align}:~\url{https://github.com/clab/fast_align}} and large length ratios are also discarded. Finally, $340$K sentence pairs are kept for training.

\begin{CJK*}{UTF8}{gbsn}
\begin{table}[!t]
\centering
\renewcommand\arraystretch{0.7}
{\scriptsize
  \begin{tabular}{m{26pt}|m{189pt}}
  \toprule
    {\bf Src.1} & There are 68 million people estimated to be in wheelchairs worldwide \\ \midrule
    {\bf OrigRef}  & $\overset{\mbox{worldwide}}{\mbox{世界上}}$ \ \  $\overset{\mbox{estimate}}{\mbox{估计}}$ \ \  $\overset{\mbox{have}}{\mbox{有}}$ \ \  $\overset{\mbox{68 million}}{\mbox{6千8百万}}$ \ \  $\overset{\mbox{wheelchair}} {\mbox{轮椅}} $ \ \  $\overset{\mbox{users}}{\mbox{使用者}}$  \\ \midrule
    {\bf MonoRef}  & $\overset{\mbox{there are 68 million,}}{\mbox{有6800万人，}}$ \ \ $\overset{\mbox{estimate}}{\mbox{估计}}$\ \  $\overset{\mbox{there are 68 million}}{\mbox{有6800万人}}$\ \ $\overset{\mbox{use}}{\mbox{使用}}$ \ \  $\overset{\mbox{wheelchair}}{\mbox{轮椅，}}$ \ \ $\overset{\mbox{worldwide}}{\mbox{在全世界范围内}}$  \\ \midrule
    {\bf Src.2}    & Who are these cousins? \\ \midrule
    {\bf OrigRef}  & $\overset{\mbox{these}}{\mbox{这些}}$ \ \ $\overset{\mbox{cousin}}{\mbox{近亲}}$ \ \ $\overset{\mbox{is what}}{\mbox{是些什么}}$ \ \  $\overset{\mbox{animal}}{\mbox{动物}}$ \\ \midrule
    {\bf MonoRef}  &  $\overset{\mbox{who}}{\mbox{谁}}$\ \  $\overset{\mbox{are}}{\mbox{是}}$\ \ $\overset{\mbox{these}}{\mbox{这些}}$\ \  $\overset{\mbox{cousin}}{\mbox{堂兄妹}}$ \\ \bottomrule
  \end{tabular}
}
\vspace{-0.2cm}
\caption{
\small{Examples of~\texttt{test-orig} and~\texttt{test-mono}.}}\vspace{-0.5cm}
  \label{table:annotation_example}
\end{table}
\end{CJK*}
\vspace{-0.2cm}
\subsection{Models}
\vspace{-0.2cm}
% 详细介绍所用到的同传模型（如常规模型、wait-k模型、MMA模型以及re-translation）以及对应的configurations To evaluate the quality-latency trade-off
% Simultaneous translation mainly adopts two strategies: streaming translation~\cite{gu2017learning,arivazhagan2019monotonic,ma2020monotonic} and re-translation~\cite{niehues2016dynamic,arivazhagan2020re_iwslt,arivazhagan2020re_icassp}.
%Streaming translation focuses on READ/WRITE decisions and re-translation strategy refreshes translation output when new source words are read in.
We employ the following three models to compare the performance of the SimulMT models on~\texttt{test-orig} and~\texttt{test-mono}:
\begin{list}{\labelitemi}{\leftmargin=1em} \vspace{-0.2cm}
\setlength{\itemsep}{0pt}
\setlength{\parsep}{0pt}
\setlength{\parskip}{0pt}
    % \item {\bf base}: Conventional sequence-to-sequence baseline model. Test-time wait-$k$~\cite{ma2018stacl} could be treated as a streaming translation baseline, and biased beam search~\cite{arivazhagan2020re_icassp} is adopted to perform a re-translation baseline. %~\footnote{Our implementation based on fairseq.}.
    % \item {\bf mix-perfix}: Following~\cite{arivazhagan2020re_iwslt}, we train the base model on the mixture of the original training set and prefix pairs.
    \item {\bf Wait-$k$}: Streaming translation models trained with fixed latency~($k$=$13$ for reported results), proposed by~\cite{ma2018stacl}. % We empirically choose $k=\{5,9,13\}$, which roughly corresponds to the low, medium, and high latency regimes.
    \item {\bf GMA}: Streaming translation models trained with an adaptive-policy strategy, proposed by~\cite{zhang2022gaussian}.% We train GMA models with $\delta=\{0.5,0.7,0.9,1,2,2.5\}$ to perform simultaneous inference with different latency.
    \item {\bf Re-trans}: Re-translation models, conventional transformer trained on the mixture of the original training set and prefix pairs~\cite{arivazhagan2020re_iwslt}, with biased beam search proposed by~\cite{arivazhagan2020re_icassp}.
\end{list} \vspace{-0.2cm}
Wait-$k$ and GMA are used to evaluate the streaming translation model, and Re-trans is for evaluating the re-translation model.
% ~\footnote{Based on the released code:~\url{https://github.com/ictnlp/GMA}.}
% ~\footnote{Based on the released code:~\url{https://github.com/facebookresearch/fairseq/tree/main/examples/simultaneous_translation}.} 
All models are implemented based on~\texttt{fairseq}~\cite{ott-etal-2019-fairseq} with the~\texttt{transformer\_iwslt\_de\_en} setting. %We set the token-level batch size as $32$k. Models are optimized by Adam optimizer~\cite{kingma2015adam} with learning rate set as $2.5$e-$4$. We train the transformer model for $30$k steps and choose the best checkpoints based on the development set.
\begin{figure*}[t!]
\centering
\begin{subfigure}{.30\textwidth}
  \centering
  \includegraphics[width=1.0\linewidth]{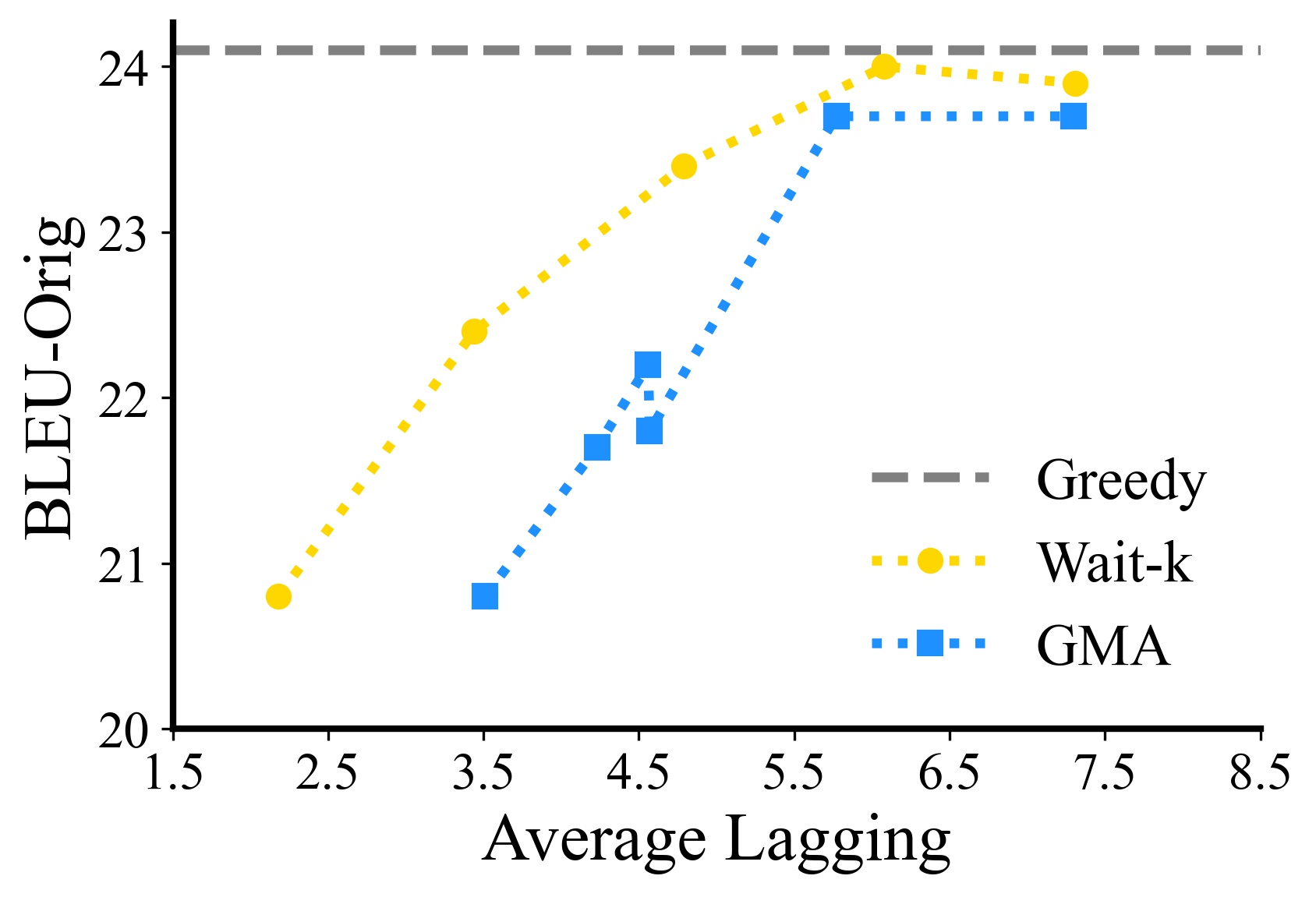}
  \caption{BLEU-Orig vs.~AL}% on the MT03 test dataset under different beam sizes}
  \label{fig:bleu_orig_al}
\end{subfigure}%
\quad
\begin{subfigure}{.30\textwidth}
  \centering
  \includegraphics[width=1.0\linewidth]{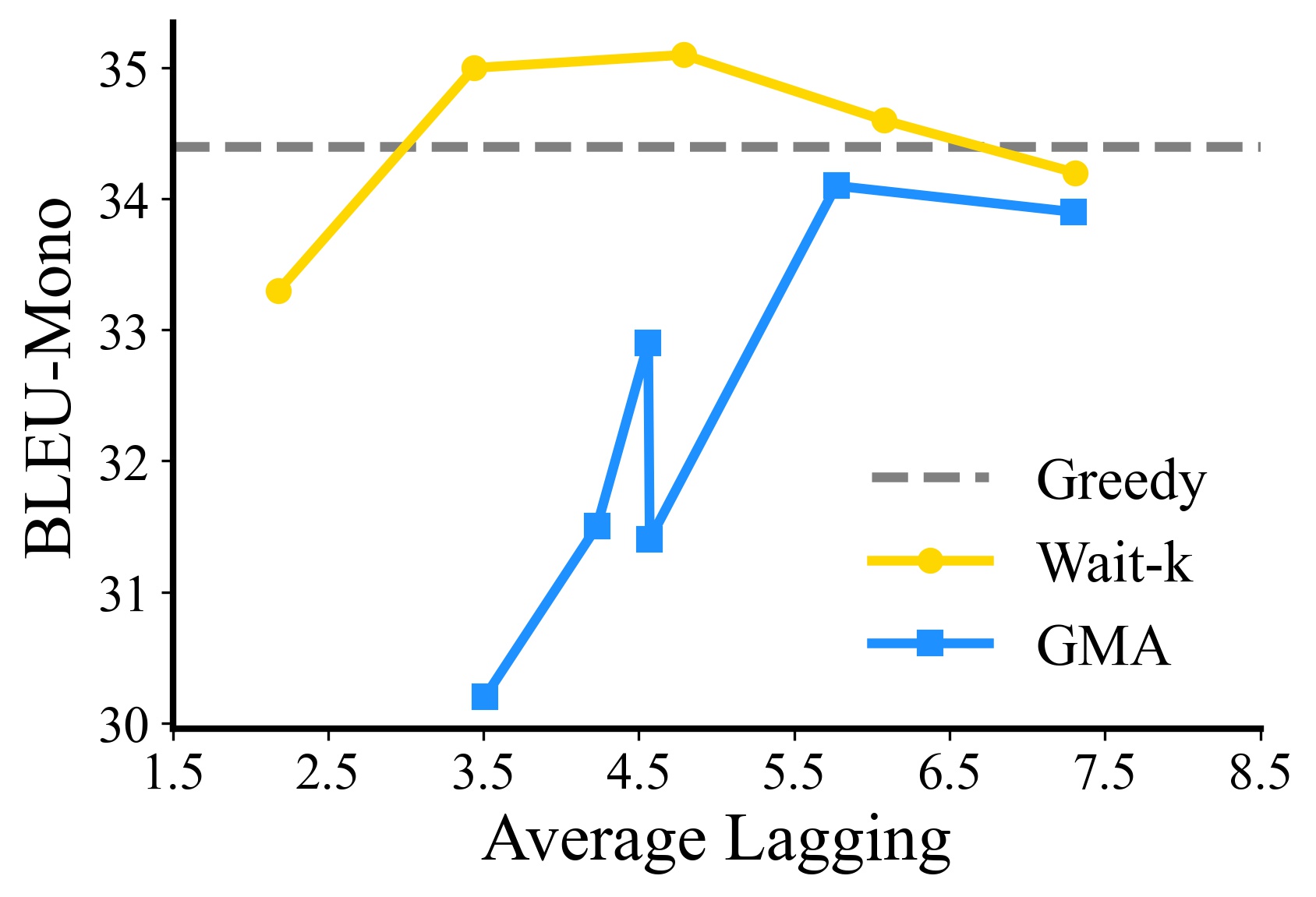}
  \caption{BLEU-Mono vs.~AL}% on the MT03 test dataset under different beam sizes}
  \label{fig:bleu_mono_al}
\end{subfigure}
\begin{subfigure}{.30\textwidth}
  \centering
  \includegraphics[width=1.0\linewidth]{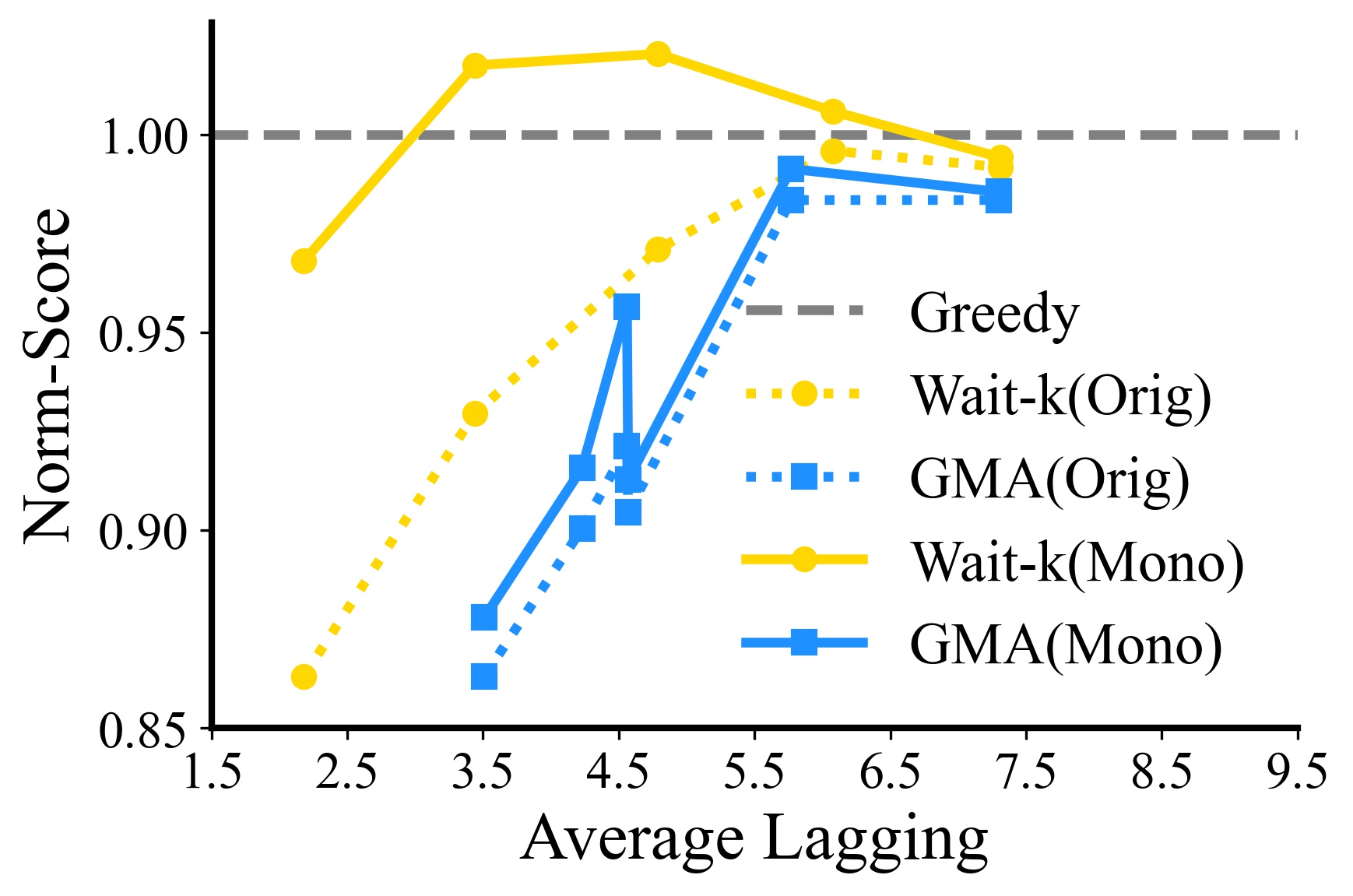}
  \caption{Norm-Score vs.~AL}% on the MT03 test dataset under different beam sizes}
  \label{fig:norm_score_al}
\end{subfigure}
\vspace{-0.2cm}
\caption{\small{BLEU-AL curves of streaming translation methods, including Wait-$k$, and GMA models. Greedy denotes the general full-sentence translation model decodes with~\texttt{beamsize} of $1$.}}\vspace{-0.5cm}
\label{figs:BLEU-AL}
\end{figure*}
\vspace{-0.2cm}
\subsection{Metrics} ~\label{sec:metrics}
% \vspace{-0.1cm}
% 详细介绍我们的各种指标
To explore the availability of our annotated test set, we conduct analysis from three aspects: quality, latency, and stability.
The BLEU~\cite{papineni2002bleu} scores on both~\texttt{test-orig} and~\texttt{test-mono} are calculated by~\texttt{SacreBleu}~\cite{post2018call}, and denoted as BLEU-Orig and BLEU-Mono, respectively.
Since the reference stream is also recorded during the annotation process, we can calculate the BLEU score of the intermediate translation, denoted as BLEU-Stream.
Following~\cite{ma2018stacl} and~\cite{arivazhagan2020re_icassp}, Average Lagging~(AL) and Normalized Erasure~(NE) are adopted to measure the latency and the stability, respectively.

\vspace{-0.2cm}
\subsection{Analytical Experiments}
\vspace{-0.2cm}
% In this part, we first verify the applicability of our annotated test set, and then compare SimulMT models on~\texttt{test-orig} and~\texttt{test-mono} in terms of quality, latency, and monotonicity.

\subsubsection{Applicability and Monotonicity}
\vspace{-0.2cm}
% 对比test-orig 和 test-mono，表格展示
% As shown in Table~\ref{table-test-compare}, the BLEU, BERTScore~\cite{zhang2019bertscore} of~\texttt{test-mono} is evaluate by~\texttt{test-orig}.
% We first randomly sample $200$ sentences from the test set, and then use the x-model to generate translations.
% For the two test sets~\texttt{test-orig} and~\texttt{test-mono}, firstly $200$ sentences are randomly sampled, then three annotators rate the acceptability of translations in the range of $[1,5]$ separately, and finally, the translation with a score of at least $3$ is considered acceptable. 

For the two test sets, ~\texttt{test-orig} and~\texttt{test-mono}, $200$ sentences are randomly sampled, then three annotators separately rate the acceptability of translations in the range of $[1,5]$, and finally, translations with a score of at least $3$ are considered acceptable. 
The average human score and average acceptability~(AP) rates on both test sets are listed in Table~\ref{table-test-compare}. It can be seen that the acceptability ratio of~\texttt{test-mono} is comparable to~\texttt{test-orig}, confirming the high quality of our annotated~\texttt{test-mono}. %, and is much higher than the AP score of the interpretation data in~\cite{zhang2021bstc}.
AA is calculated according to Equation~\ref{AA-AR}, reflecting the monotonicity of the reference translation, and the smaller value means the better monotonicity.
The AL of~\texttt{test-orig} is counted by the number of words in the source sentence, and the AL of~\texttt{test-mono} is calculated based on the number of waiting words per WRITE action during the annotation process.
Both AA and AL scores are averaged over the test set.
% The simulated-AL is calculated by the ideal delay of each target token, which is obtained by alignments. And simulate-AL reflects the ideal minimum latency of the reference sentence.
Table~\ref{table-test-compare} shows that~\texttt{test-mono} has lower AL and AA scores than~\texttt{test-orig}, which indicates that~\texttt{test-mono} is an online annotated test set and has better monotonicity.
%Low AL of ~\texttt{test-mono} proves that it is constructed online, and lower AA of ~\texttt{test-mono} compared to ~\texttt{test-orig} indicate less long-distance reordering phenomena in~\texttt{test-mono}.
In conclusion,~\texttt{test-mono} is of high quality and more monotonic.
\vspace{-0.3cm}
% Translation interval is a special value for annotation streams. For \verb|test-naive| translation is infinite because translation is performed on full sentences. 
% In annotation streams, we count the average number of waited words between each time the annotator gives a new translation as a translation interval. We could say the annotator performs translation on every 3-4 words on average.
\begin{table}
\centering
\renewcommand\arraystretch{1.2}
{\small
\begin{tabular}{c|cc}
\toprule
\textbf{Metrics} & \textbf{\texttt{test-orig}} & \textbf{\texttt{test-mono}} \\
\hline
% BLEU $\uparrow$ & - & $26.4$  \\
% BERTScore $\uparrow$ & - & $80.91$  \\
Human Score $\uparrow$ & $4.24$ & $4.08$ \\
AP $\uparrow$ & $92.7\%$ & $89.7\%$ \\
AA $\downarrow$ & $1.47$ & $0.77$ \\
% \small{Translation interval} & \small{inf} & \small{3.81} \\
AL $\downarrow$ & $15.65$ & $2.71$ \\
% Simulate-AL & $2.08$ & $1.53$ \\
\bottomrule
\end{tabular}
\vspace{-0.2cm}
}
\caption{\label{table-test-compare}
\small{Comparison between~\texttt{test-orig} and~\texttt{test-mono}.}
%  AA is metric about monotonicity, lower means more monotonic. AL is a metric about latency, smaller means lower latency.
}\vspace{-0.5cm}
\end{table}

\begin{figure*}[t!]
\centering
\begin{subfigure}{.30\textwidth}
  \centering
  \includegraphics[width=1.0\linewidth]{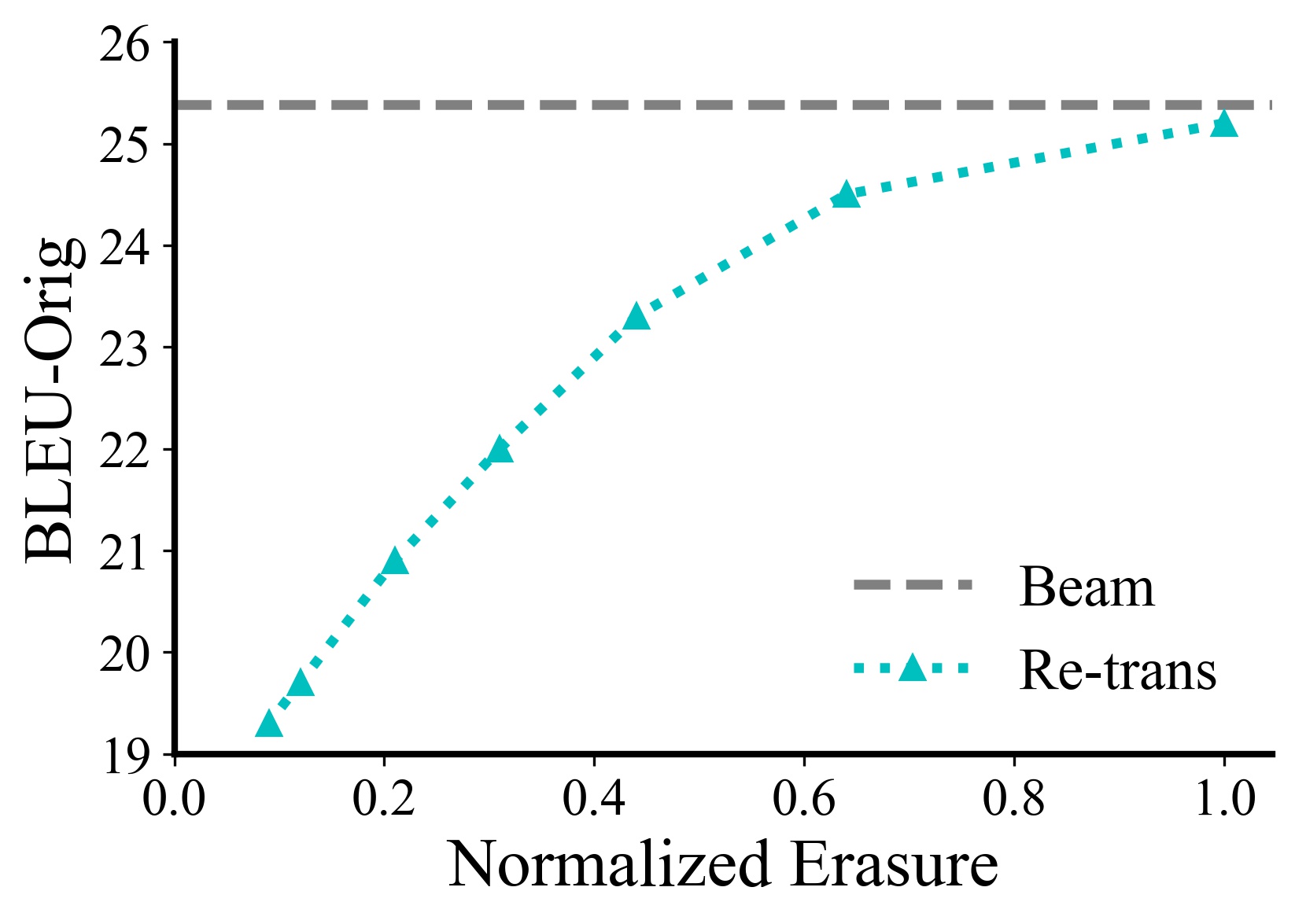}
  \caption{BLEU-Orig vs.~NE}% on the MT03 test dataset under different beam sizes}
  \label{fig:bleu-orig-ne}
\end{subfigure}%
\quad
\begin{subfigure}{.30\textwidth}
  \centering
  \includegraphics[width=1.0\linewidth]{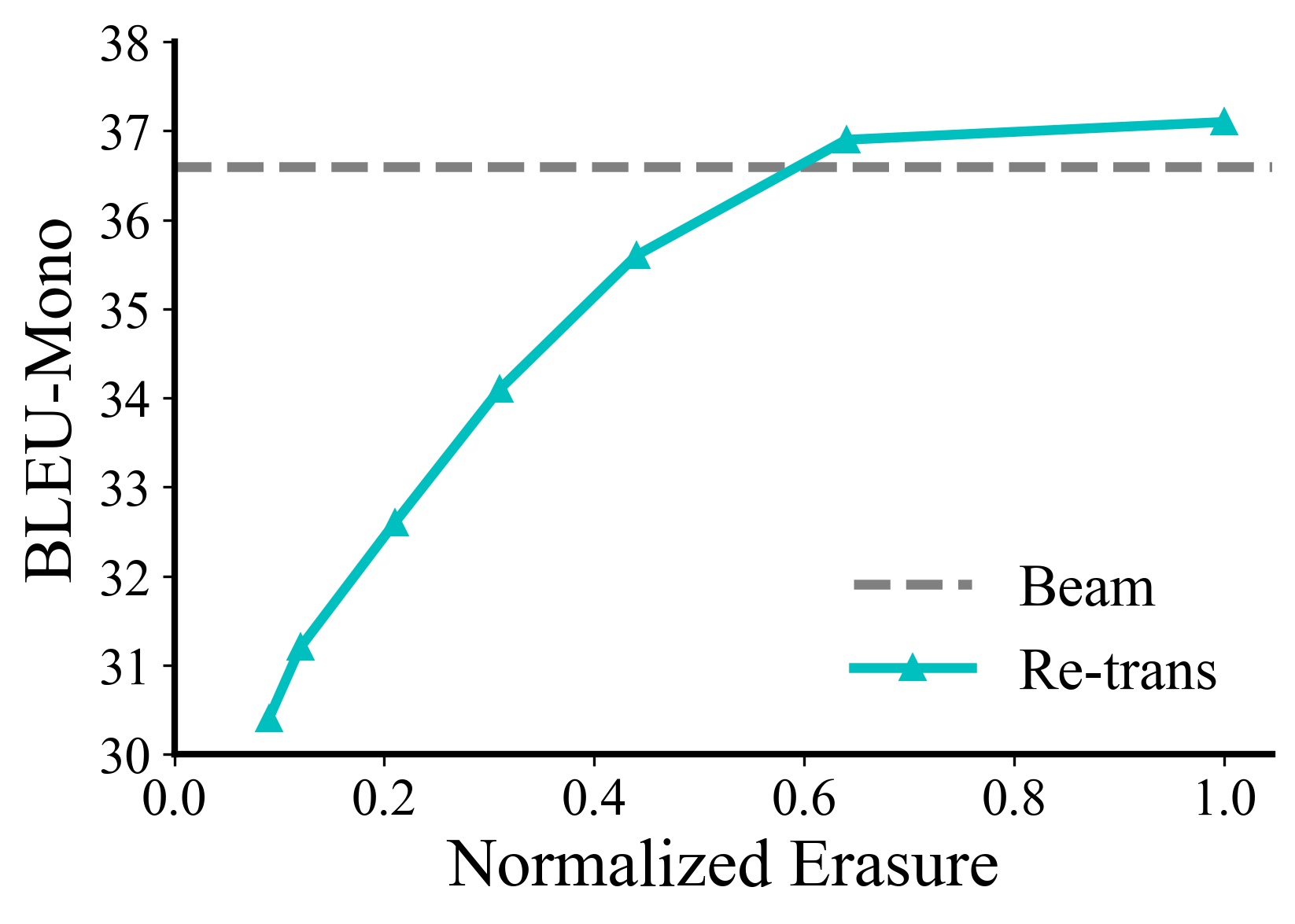}
  \caption{BLEU-Mono vs.~NE}% on the MT03 test dataset under different beam sizes}
  \label{fig:bleu-mono-ne}
\end{subfigure}
\begin{subfigure}{.30\textwidth}
  \centering
  \includegraphics[width=1.0\linewidth]{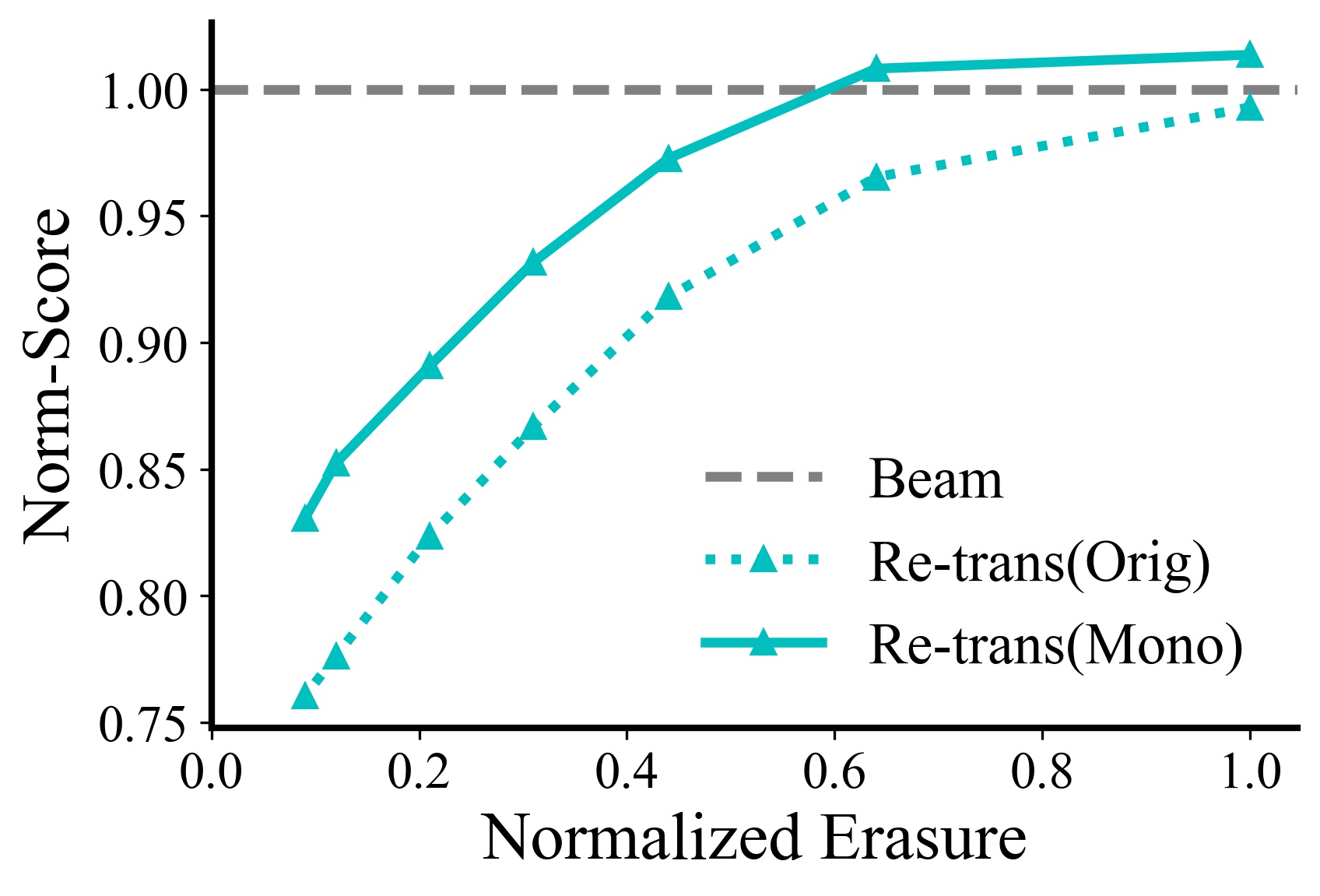}
  \caption{Norm-Score vs.~NE}% on the MT03 test dataset under different beam sizes}
  \label{fig:norm-score-ne}
\end{subfigure}
\vspace{-0.2cm}
\caption{\small{BLEU-NE curves of re-translation strategy. Lower NE means better stability. Beam denotes the general full-sentence translation.}} \vspace{-0.3cm}
%  model decodes with~\texttt{beamsize} of $5$.
\label{figs:BLEU-NE}
\end{figure*}

\begin{figure*}[t!]
\centering
\begin{subfigure}{.30\textwidth}
  \centering
  \includegraphics[width=1.0\linewidth]{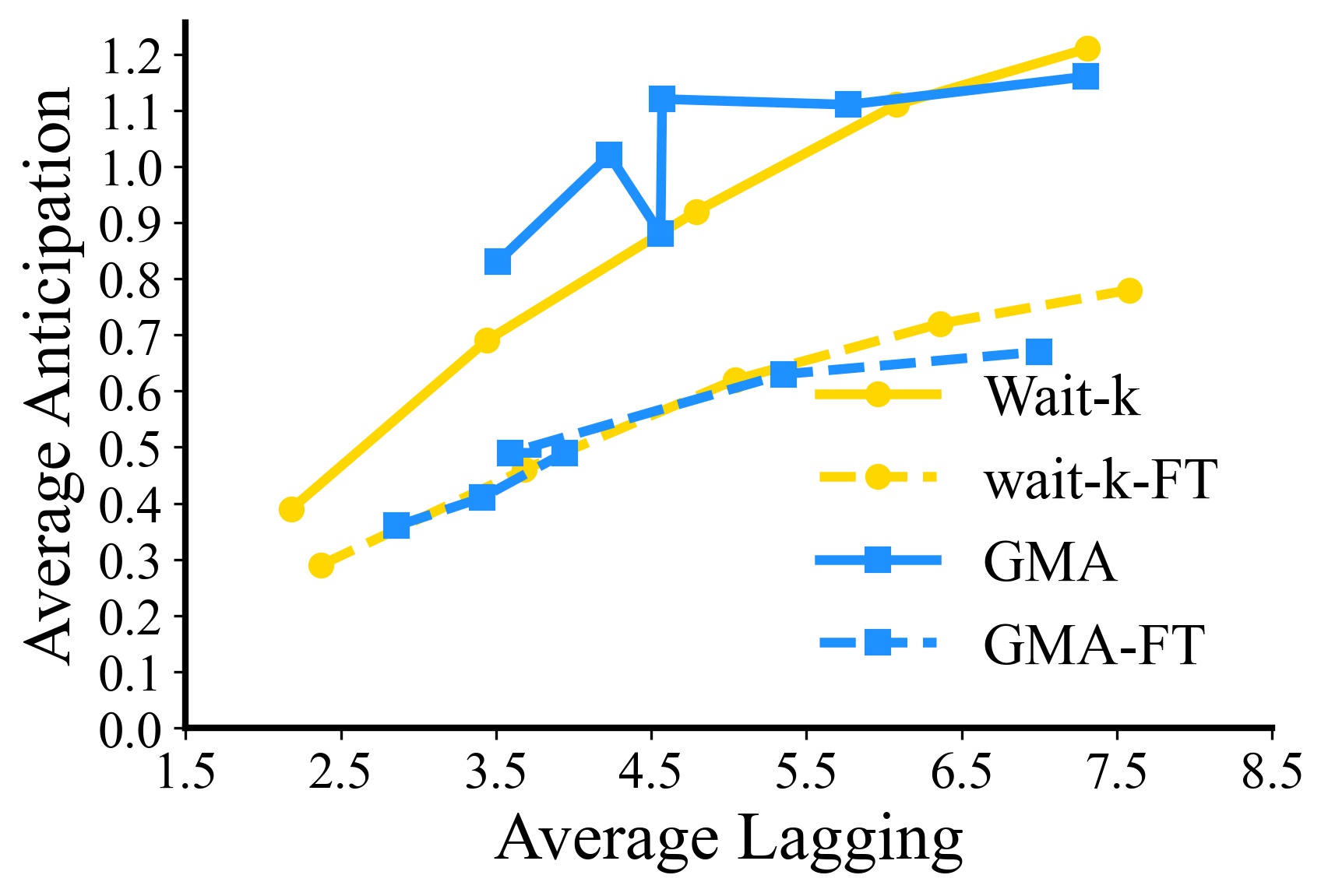}
  \caption{AA vs. AL}% on the MT03 test dataset under different beam sizes}
  \label{fig:ft_aa_al}
\end{subfigure}
\begin{subfigure}{.30\textwidth}
  \centering
  \includegraphics[width=1.0\linewidth]{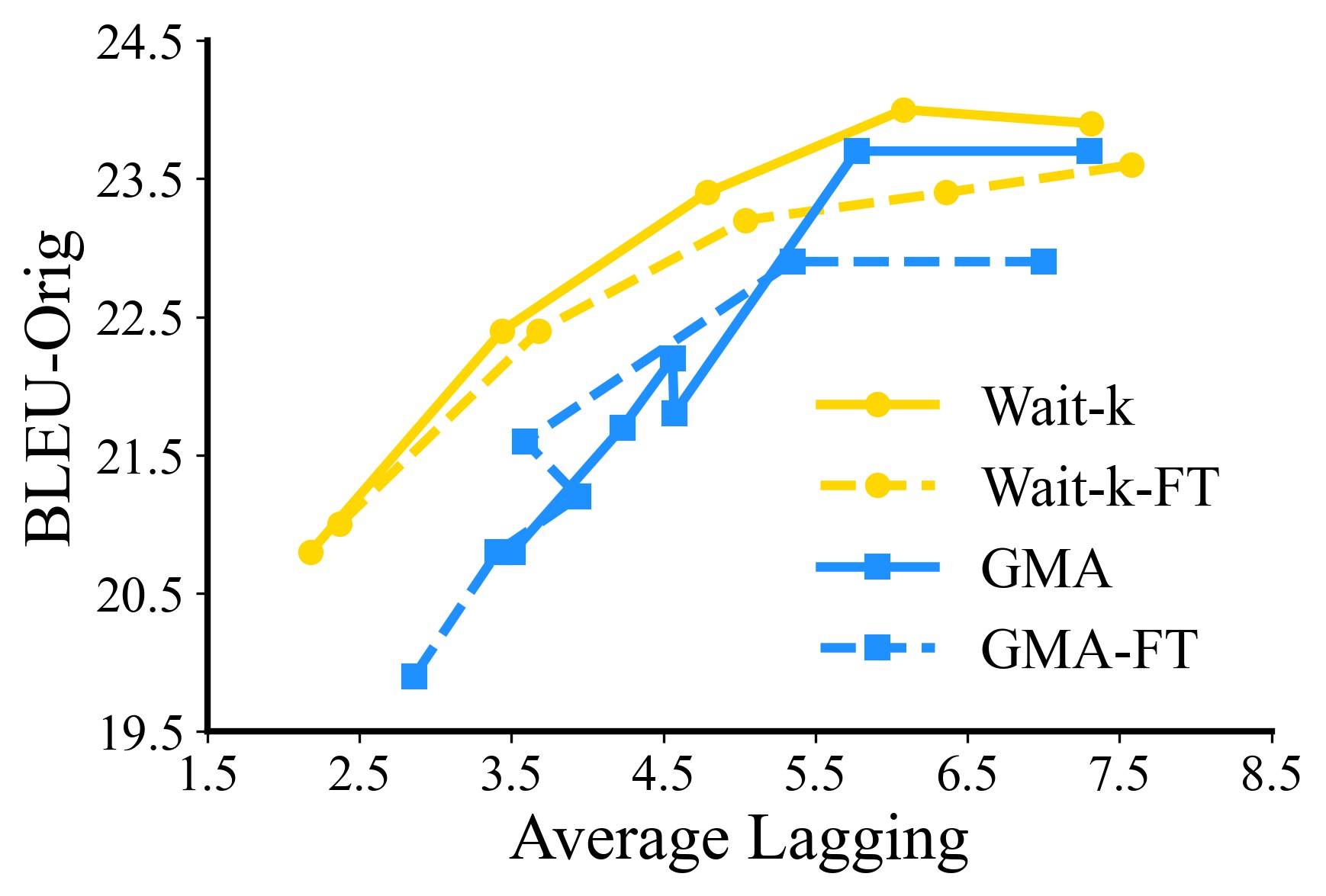}
  \caption{BLEU-Orig vs.~AL}% on the MT03 test dataset under different beam sizes}
  \label{fig:ft_bleu_orig_al}
\end{subfigure}%
\quad
\begin{subfigure}{.30\textwidth}
  \centering
  \includegraphics[width=1.0\linewidth]{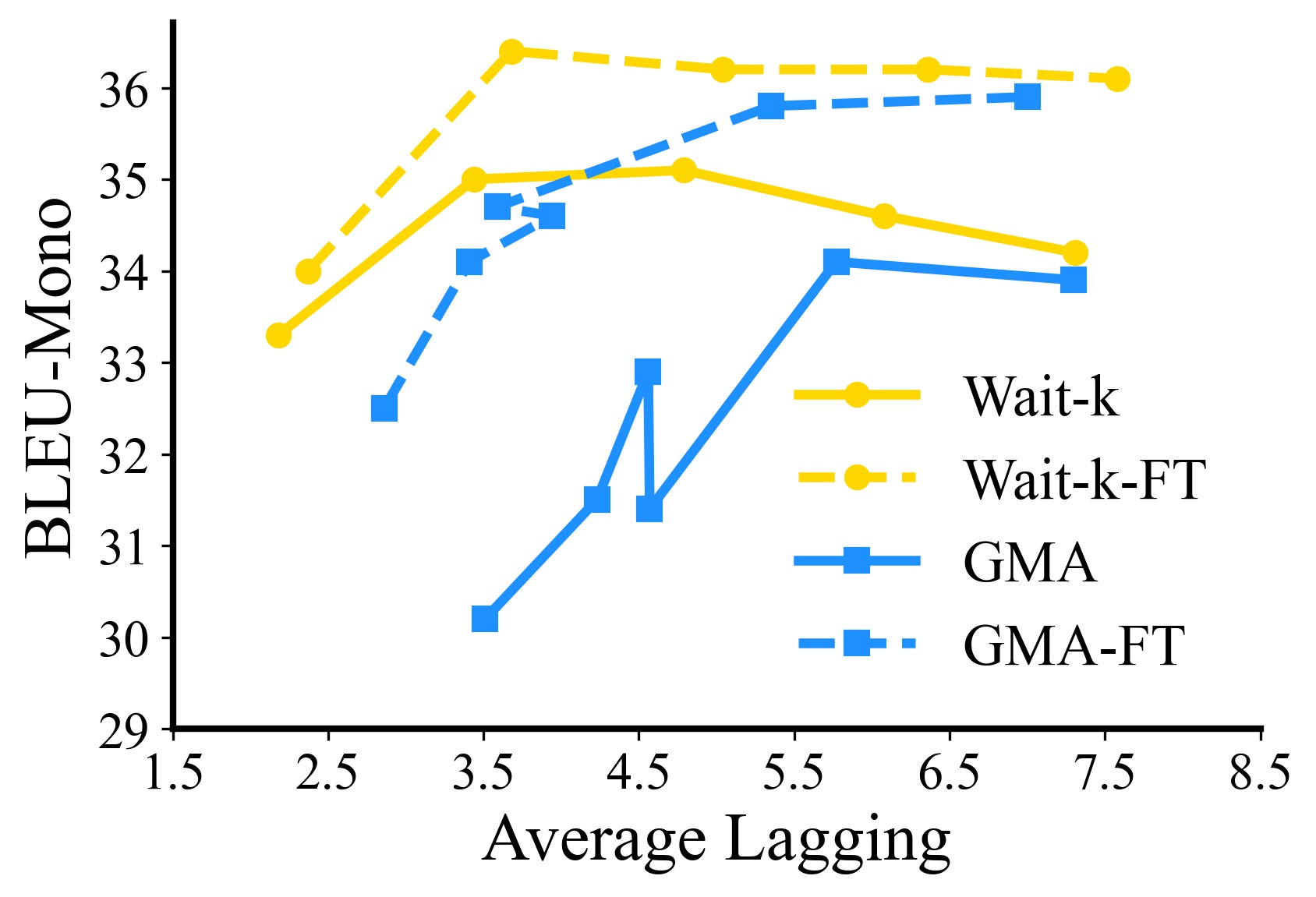}
  \caption{BLEU-Mono vs.~AL}% on the MT03 test dataset under different beam sizes}
  \label{fig:ft_bleu_mono_al}
\end{subfigure}
\vspace{-0.2cm}
\caption{\small{Finetuning results of Wait-$k$ and GMA models. Wait-$k$-FT and GMA-FT denote the finetuned models. The lower AA means better monotonicity. }}\vspace{-0.5cm}
\label{figs:FT-BLEU-AA-AL}
\end{figure*}

\vspace{-0.3cm}
\subsubsection{Quality and Latency}
\label{sec:quality_latency}
\vspace{-0.2cm}
We leverage the BLEU-AL curves to show the trade-off between the quality and latency of the SimulMT model. As shown in Figure~\ref{figs:BLEU-AL}, BLEU-Orig, BLEU-Mono, and normalized scores on the test sets are calculated separately. Note that, as explained in Section~\ref{subsec:datasets}, the BLEU scores of SimulMT models on~\texttt{test-mono} are higher than those on~\texttt{test-orig}.
    
In Figures~\ref{fig:bleu_orig_al} and~\ref{fig:bleu_mono_al}, the Wait-$k$ model performs better than the GMA model when evaluated on~\texttt{test-orig} and~\texttt{test-mono}, the Wait-$k$ model has a higher BLEU score in each AL regime. The GMA model performs poorly, possibly due to the reordering in English-Chinese parallel data, which makes it difficult to learn the best READ/WRITE path.
The quality drop of the Wait-$k$ model in high-latency regime can be observed in Figure~\ref{fig:bleu_mono_al}, which is caused by the decrease in data monotonicity as more source information is read.
% For the same system, the values of BLEU-Mono are higher than BLEU-Orig, which may be because~\texttt{test-mono} has better royalty than~\texttt{test-orig} as described in Section~\ref{subsec:datasets}. So the sentence-level translation generated by SimulMT models obtains higher scores.
For comparability across different test sets, we compute the Norm-Score to normalize the BLEU scores across different test sets.
For each test set, the BLEU score of the full-sentence translation is regarded as the base value (marked as grey dash lines in Figures~\ref{fig:bleu_orig_al} and~\ref{fig:bleu_mono_al}), and the BLEU scores of SimulMT models are divided by the base value to get the Norm-Score.
As shown in Figure~\ref{fig:norm_score_al}, in the low-latency regime, the Wait-$k$ model performs significantly better on~\texttt{test-mono} than on~\texttt{test-orig}.
%When compared by Norm-Score in Figure~\ref{fig:norm_score_al}, in the low-latency regime, scores of \verb|test-mono| represented by solid lines are much higher than scores of \verb|test-orig| represented by dash lines.
This significant improvement indicates that the translation quality at low latency is seriously underestimated by~\texttt{test-orig}.
It can be concluded that \verb|test-mono| provides evaluation results more consistent with human evaluation, without underestimation caused by long-distance reordering.

\vspace{-0.3cm}
\subsubsection{Quality and Stability}
\vspace{-0.2cm}
The results of the re-translation methods are shown in Figure~\ref{figs:BLEU-NE}. We draw BLEU-NE curves to show the quality-stability trade-off. Both BLEU-Orig and BLEU-Mono decrease as NE becomes lower.
We apply the same normalization method to compare the BLEU scores on ~\texttt{test-orig} and~\texttt{test-mono}.
% Stability improves when prefix data is mixed in the training corpus, which is consistent with the results reported in~\cite{arivazhagan2020re_iwslt}.
As shown in Figure~\ref{fig:norm-score-ne}, the normalized scores on ~\texttt{test-mono} are shown in a solid line and scores on ~\texttt{test-orig} are shown in a dash line. The solid line achieves higher scores, especially in the few-flicker regime, which is consistent with the results of quality and latency analysis in Section~\ref{sec:quality_latency}.
The results of the re-translation methods provide further evidence that our~\texttt{test-mono} is more consistent with human evaluation.

\begin{figure}[!t]
    \centering
    \includegraphics[scale=0.33]{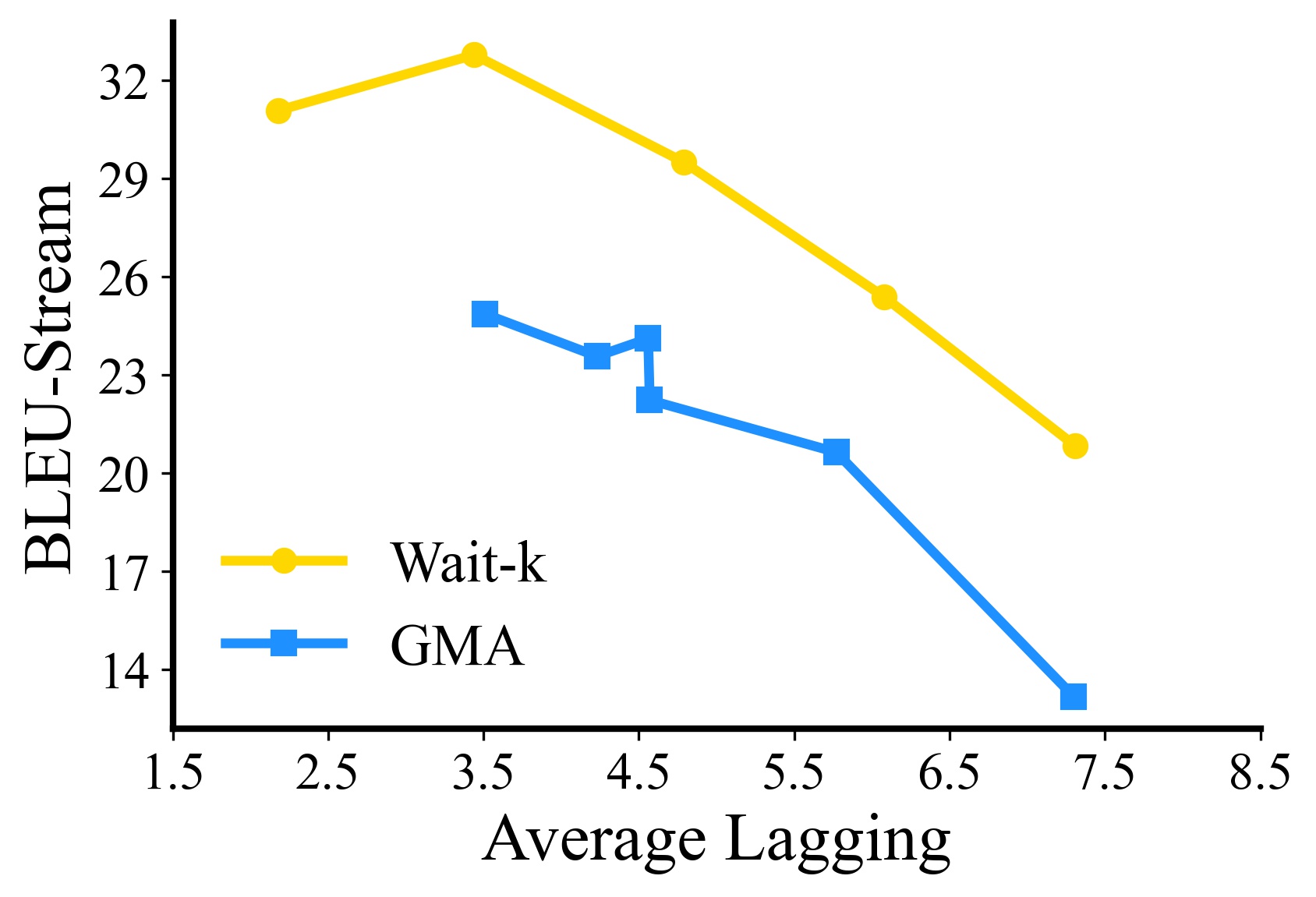}
    \vspace{-0.3cm}
    \caption{
    \small{BLEU-steam vs.AL of streaming translation models.} 
    }\vspace{-0.5cm}
    \label{figs:bleu-stream-al}
\end{figure}

\vspace{-0.3cm}
\subsubsection{Steaming Evaluation}
\vspace{-0.2cm}
The BLEU-Stream evaluation is shown in Figure~\ref{figs:bleu-stream-al}. The BLEU-Stream scores reach the highest value when AL is about $3.5$ then decrease as latency becomes higher. This is different from other BLEU-AL curves because the target streams may have many blank or short translation in high-latency regime. The BLEU-Stream score may provide us with a reference latency regime, which is close to the delay of manual annotation.

\vspace{-0.3cm}
\subsubsection{Finetuning}
\vspace{-0.2cm}
% finetune method

% after finetuning, aa decrease, BLEU increase
To enhance the monotonicity of SimulMT models, we select $42,000$ sentence pairs with no anticipation from the original training corpus, denoted as the monotonic corpus. On the monotonic corpus, we finetune the Wait-$k$ and GMA models for $2,000$ steps. 
And we calculate the AA scores~(Equation~\ref{AA-AR}) of the hypotheses generated by models for monotonicity evaluation. 
Figure~\ref{fig:ft_aa_al} shows that the AA scores of the Wait-$k$ and GMA models are both significantly lower after finetuning, meaning the monotonicity improvement of SimulMT models when optimized by the monotonic corpus. 

Figure~\ref{fig:ft_bleu_orig_al} and Figure~\ref{fig:ft_bleu_mono_al} present the impact of finetuning to BLEU scores. After finetuning, for both Wait-$k$ and GMA models, the BLEU-Orig grows a little bit in low-latency regime and decreases in higher-latency regime. The BLEU-Mono of the Wait-$k$ and GMA models, on the other hand, improves dramatically. In particular, when evaluated by BLEU-Mono, the GMA model is improved by more than 3 points, and the latency gets lower at each $\delta$ setting~\cite{zhang2022gaussian}. This notable improvement may suggest that the monotonic corpus is much easier for the GMA model to learn the READ/WRITE strategy. 
As the finetuning benefits the monotonicity of SimulMT models, our ~\texttt{test-mono} can better reflect this improvement because of its better monotonicity.
So the~\texttt{test-mono} performs better in evaluating the monotonicity of SimulMT models. 
\vspace{-0.2cm}

\section{Conclusion}
\vspace{-0.2cm}
We design a streaming annotation method to annotate a monotonic test set based on the MuST-C English-Chinese test set. Human evaluation and experiments prove that our SiMuST-C is of high quality and has better monotonicity.
Besides, the automatically extracted monotonic training set can help SimulMT models generate monotonic translations and also significantly improve the model's performance. % on our annotated test set.
Overall, our annotated monotonic test set is more suitable for the evaluation of English-Chinese simultaneous machine translation.
% But whether monotonicity is equally important to simultaneous translation between other languages is still worthy of research and exploration.
% We speculate that the SimulMT models could be further improved with more manually annotated monotonic training sets, but the cost of manual annotation is relatively high, and we will explore cheaper data acquisition methods in future work. This paper provides an alternative direction for the NLP community.
% 不写future work
\vspace{-0.3cm}
\section{Acknowledgments}
\vspace{-0.2cm}
This work is supported in part by the National Key R\&D Program of China (No. 2020AAA0106600)

\vfill\pagebreak
\bibliographystyle{IEEEbib}
\bibliography{strings,refs}

\clearpage

\end{document}